\documentclass[10pt,twocolumn,letterpaper]{article}

\usepackage{wacv}
\usepackage{times}
\usepackage{epsfig}
\usepackage{graphicx}
\usepackage{amsmath}
\usepackage{amssymb}

\usepackage{booktabs}
\usepackage{subcaption}

\def\wacvPaperID{381} 

\wacvfinalcopy 

\ifwacvfinal
\def\assignedStartPage{1} 
\fi

\ifwacvfinal
\usepackage[breaklinks=true,bookmarks=false]{hyperref}
\else
\usepackage[pagebackref=true,breaklinks=true,colorlinks,bookmarks=false]{hyperref}
\fi

\ifwacvfinal
\else
\fi

\begin{document}

\title{Learning Maritime Obstacle Detection from Weak Annotations by Scaffolding}

\author{
Lojze Žust \qquad Matej Kristan\\
University of Ljubljana, Faculty of Computer and Information Science\\
Ljubljana, Slovenia\\
{\tt\small \{lojze.zust,matej.kristan\}@fri.uni-lj.si}
}

\maketitle

\begin{abstract}
Coastal water autonomous boats rely on robust perception methods for obstacle detection and timely collision avoidance. The current state-of-the-art is based on deep segmentation networks trained on large datasets. Per-pixel ground truth labeling of such datasets, however, is labor-intensive and expensive.
We observe that far less information is required for practical obstacle avoidance -- the location of water edge on static obstacles like shore and approximate location and bounds of dynamic obstacles in the water is sufficient to plan a reaction.
We propose a new scaffolding learning regime (SLR) that allows training obstacle detection segmentation networks only from such weak annotations, thus significantly reducing the cost of ground-truth labeling.
Experiments show that maritime obstacle segmentation networks trained using SLR substantially outperform the same networks trained with dense ground truth labels. Thus accuracy is not sacrificed for labelling simplicity but is in fact improved, which is a remarkable result.


\end{abstract}

\section{Introduction}

Autonomous boats are an emerging research area with significant application potential in cross-ocean cargo shipping, coastal environment control and man-made structure inspection.
Their autonomy crucially depends on perception capability, which is particularly challenging in environments like coastal waters, marinas, city canals and rivers. There, the appearance of the navigable area -- the water -- significantly varies with weather conditions, surroundings, contains mirrored reflections of land and sun glitter. The appearance of potential obstacles is equally broad. Obstacles may be static (\eg shore and piers) or dynamic (\eg boats, swimmers, debris, buoys).

\begin{figure}[t]
\centering
\includegraphics[width=\linewidth]{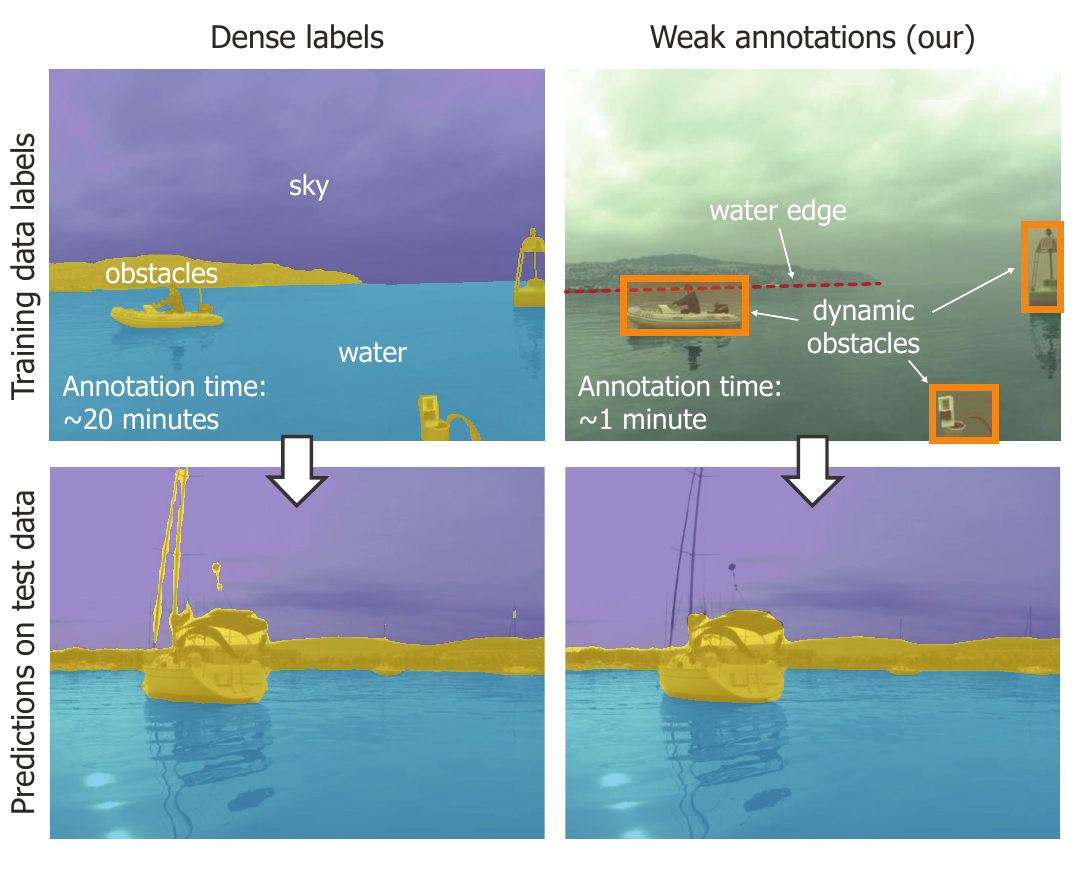}
\caption{The proposed scaffolding learning regime allows training a segmentation network using weak annotations (upper right) without hampering the segmentation quality in aspects important for the obstacle detection task (bottom row), thus avoiding the need for laborious dense per-pixel labeling of training images (upper left).}
\label{fig:annotations}
\end{figure}

The current state-of-the-art approaches in maritime obstacle detection harness the power of deep models~\cite{Bovcon2020WaSR,Yang2019Surface,Steccanella2020} to capture the significant appearance variability and segment the onboard-captured RGB images into water, sky and obstacle classes (Figure~\ref{fig:annotations}). These methods are currently trained on carefully manually labeled segmentation datasets~\cite{Bovcon2019Mastr}, the annotation of which is laborious, error-prone and costly.

Furthermore, obstacle avoidance does not require equally high segmentation accuracy in all image regions.
For example, detecting the boundary between water and shore is crucial for collision prevention, while accurate segmentation of the shore-sky boundary is meaningless from the perspective of obstacle avoidance. Similarly, missing a few edge pixels on a swimmer will not cause a collision. But falsely classifying isolated patches of water pixels in front of the boat as obstacles will detrimentally affect the control, causing frequent unnecessary stops. 
This is reflected in recent marine benchmarks~\cite{Bovcon2020MODS}, where performance is evaluated in terms of dynamic obstacle detection and water-edge estimation accuracy, while the segmentation accuracy beyond the water-obstacle boundary is ignored.

The information meaningful for obstacle avoidance can thus be encoded by water-obstacle boundaries of static obstacles (we refer to these simply as \textit{water edges} in the following) and by bounding boxes to denote the location and extent of dynamic obstacles. 
This raises the question of whether this simpler task-oriented annotation procedure, which takes less than a minute per image, can be leveraged for training segmentation networks, replacing labour-intensive per-pixel labeling, which typically takes over 20 minutes per image~\cite{Bovcon2019Mastr}. 

We demonstrate that weak annotations can indeed form strong constraints on the pixel labels. For example, all pixels below a water edge can be labelled as water except those within obstacle bounding boxes. Similarly, pixels above the horizon, which can be estimated from the on-board inertial measurement unit (IMU) measurements~\cite{Bovcon2018Obstacle}, may be obstacles or sky, but not water.
These constraints can be extrapolated to generate partial labels, which are dense in some regions while remaining ambiguous regions can be ignored during training.

A typical encoder-decoder network trained from such partial (\ie incomplete) labels cannot be expected to reach the desired accuracy since the labels are not sufficient to learn domain-specific context priors in the decoder to consolidate the encoder features and refine the segmentation. However, we hypothesize that these labels are sufficient to learn powerful domain-adapted visual features in the encoder. The learned features can in turn be leveraged for estimating the labels of ambiguous regions of the partial labels.

Based on these observations, we propose a scaffolding training regime (SLR) which is our main contribution. It avoids the need for manually annotated per-pixel segmentation labels, while simultaneously directing the network to focus on learning segmentation relevant for the downstream obstacle avoidance task. It utilizes the constraints from the ground-truth water edge, obstacle bounding boxes and estimated horizon to improve the encoder features. Features are in turn used to refine the unlabeled regions of the constraints-generated partial labels. 
Experimental results on the currently most challenging maritime obstacle detection dataset~\cite{Bovcon2020MODS} show that models trained using SLR outperform models classically trained from full dense annotations, which is a remarkable result. 
To the best of our knowledge, this is the first method for training obstacle detection from weak annotations in the marine domain which surpasses fully supervised training from dense labels.

\begin{figure*}
\centering
\includegraphics[width=\linewidth]{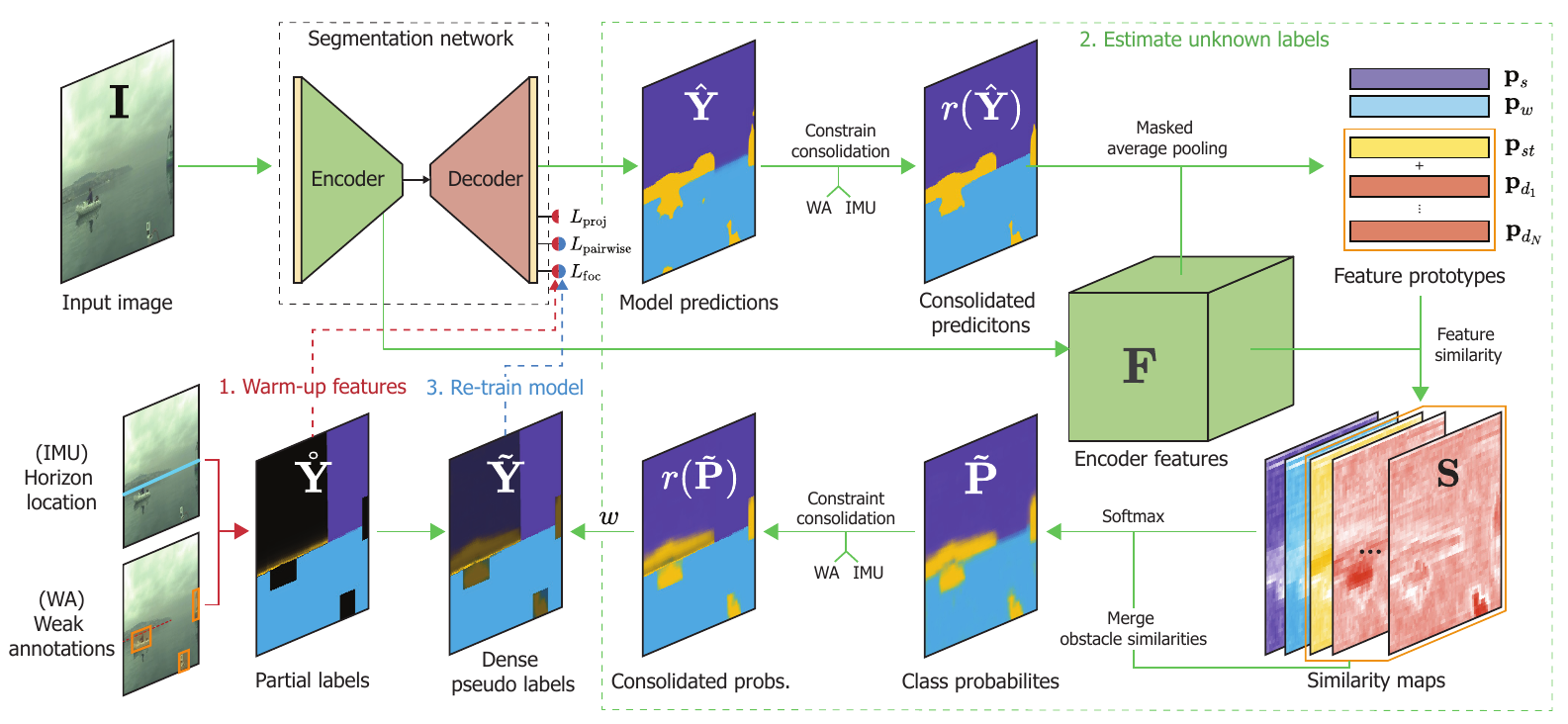}
\caption{The proposed scaffolding approach SLR is comprised of three steps, denoted by red, green and blue arrows. 
First the model is warmed-up using constraint-generated partial labels (red). 
Dense pseudo labels are estimated by completing the unknown regions in partial labels by softmax over similarity with instance prototypes computed from the warmed-up encoder
(green). 
Finally, the network is re-trained by the dense pseudo labels (blue).}
\label{fig:introspection}
\end{figure*}

\section{Related work}\label{sec:related_work}


In this section, we review the most recent works in maritime obstacle detection (Section~\ref{sec:related/detection}) and overview label-efficient training methods for deep visual models (Section~\ref{sec:related/efficient}).

\subsection{Maritime obstacle detection}\label{sec:related/detection}

Similarly to the autonomous ground vehicles (AGV) domain, perception in marine robotics has been dominated by deep convolutional neural networks (CNNs) in recent years. Several works applied~\cite{Lee2018Image,Moosbauer2019Benchmark,Yang2019Surface} or slightly extended~\cite{Ma2020Convolutional} general object detection models~\cite{Ren2017Faster,He2020Mask} from other domains to the aquatic domain for the detection of different types of ships. While these approaches achieve state-of-the-art results on the specific task of ship detection, they do not address general dynamic obstacles (\eg buoys, swimmers, debris) and static obstacles (\eg shore, land, piers, mooring posts). The latter are especially problematic for obstacle detection methods as they cannot be appropriately interpreted as objects.

State-of-the-art methods thus address the general obstacle detection by semantic segmentation. Recent works~\cite{Bovcon2019Mastr,Cane2019Evaluating} apply well-established semantic segmentation models from the AGV domain~\cite{Zhao2017Pyramid,Chen2017Rethinking} to the marine domain and outperform previous hand-crafted methods~\cite{Kristan2016Fast,Bovcon2018Stereo,Bovcon2018Obstacle}, but still perform poorly on reflections and small obstacles. Several works thus improve performance by maritime-domain-specific modifications of deep architectures~\cite{Kim2019Vision,Steccanella2020}. Notably, \cite{Bovcon2020WaSR} estimate the horizon location in the image from an onboard IMU and propose an encoder-decoder architecture that fuses the inertial information with the RGB image for accurate semantic segmentation. This architecture significantly improves obstacle detection and represents the current state-of-the-art in the field.

\subsection{Reducing the annotation effort}\label{sec:related/efficient}

One of the bottlenecks in the development of deep segmentation models is their reliance on large amounts of accurately labeled training data. 
Unlike in the more mature AGV domain, only a few (relatively small scale) segmentation datasets exist in the autonomous boats domain~\cite{Bovcon2019Mastr,Moosbauer2019Benchmark}, which hampers development. In contrast, datasets with weaker forms of annotation (\eg bounding boxes) are much more common~\cite{Kristan2016Fast,Prasad2017Video,Patino2017,Bovcon2018Stereo}.

Recently methods for annotation-efficient training have emerged, which focus on achieving the best model performance for the least amount of manual annotation effort.
Semi-supervised methods~\cite{Souly2017Semi,Hung2019Adversarial,Kalluri2019Universal,Mittal2021Semi} utilize a small set of images with dense annotations and a large set of unlabeled images to capture the diversity of the data and thus improve segmentation performance. However, as the annotation effort is mainly reduced by labeling fewer images, such methods are more sensitive to labeling errors and can only capture limited visual variation of open-ended classes like obstacles.

Alternatively, instead of reducing the number of labeled images, weakly supervised methods reduce the annotation effort by using weaker forms of labels. In the segmentation domain, approaches like scribbles~\cite{Lin2016ScribbleSup,Vernaza2017Learning,Zhang2020Weakly}, point annotations~\cite{Bearman2016What,Maninis2018Deep,Akiva2020Finding} and image-level labels~\cite{Wei2018Dilated,Huang2018Seeded,Ahn2018Affinity,Wang2020Self} have been explored. 

Among them, bounding boxes are relatively easy to annotate, while providing an informative constraint on the object bounds. Their ubiquitous presence across various perception datasets and domains makes them an ideal candidate for weakly supervised learning. Thus several works investigated bounding boxes as a viable constraint for tasks such as semantic segmentation~\cite{Dai2015BoxSup,Kulharia2020Box2Seg}, instance segmentation~\cite{Khoreva2017Simple,Hsu2019Weakly,Tian2020BoxInst} and video object segmentation~\cite{Bhat2020Learning,Zhao2021Generating}.

These approaches mainly focus on the segmentation of foreground objects that can be well approximated by bounding boxes. However, obstacle detection for AGVs or unmanned surface vehicles (USVs) also requires accurate estimation of boundaries between background classes (\eg water edge, road boundary), where such approaches cannot be applied. In contrast, our proposed scaffolding method is able to efficiently learn background classes boundaries as well, utilizing the information from water-edge annotations and the estimated horizon location.


\section{Learning to segment by scaffolding}

At a high level, our scaffolding learning regime (SLR) gradually improves the trained model by iterating between improving the network parameters (\ie training) and improving the dense pseudo labels. In practice, the learning is unfolded into three steps (Figure~\ref{fig:introspection}). In the first step (\ie feature warm-up) the network is trained using partial labels -- weak annotations and domain constraints are used to label parts of the image, while other regions remain unlabeled.
In the second step, dense pseudo labels are generated by estimating the most likely values of labels for unknown regions in partial labels from domain-specific visual features learned by the network during the warm-up phase. Finally, the network is re-trained with the estimated dense pseudo labels. These three steps are detailed in Section~\ref{sec:warmup}, Section~\ref{sec:dense_labels} and Section~\ref{sec:retraining}.

\subsection{Feature warm-up}\label{sec:warmup}


The purpose of the feature warm-up step is to learn domain-specific encoder features and initial decoder predictions. To achieve this, we supervise the network in a weakly supervised way. Specifically, combining domain knowledge and weak annotations, we can label certain regions of an input image $\mathbf{I} \in \mathbb{R}^{W \times H \times 3}$ with high confidence, while others remain unlabeled, producing partial labels $\mathring{\mathbf{Y}} \in [0,1]^{W \times H \times 3}$ (Section~\ref{sec:warmup/labels}), which can be used to supervise the model through a standard segmentation loss (Section~\ref{sec:warmup/training}).

\subsubsection{Partial labels from weak annotations}\label{sec:warmup/labels}

\begin{figure}
\centering
\includegraphics[width=\linewidth]{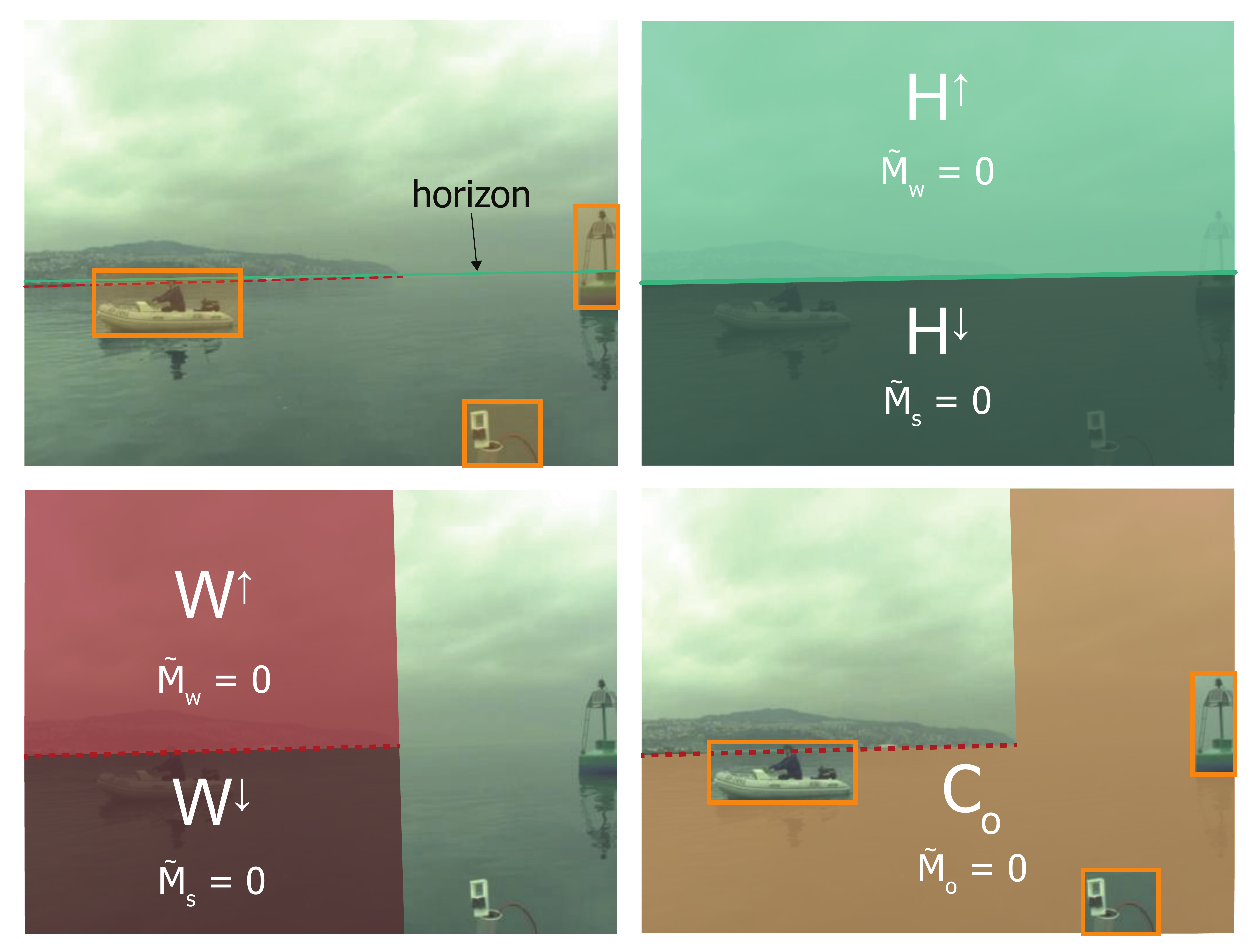}
\caption{Domain-specific constraints are computed from weak annotations and the estimated horizon (upper left).
The horizon and the water edge annotations restrict occurrence of water and sky labels (upper right, lower left), while bounding boxes restrict obstacle occurrence (lower right).}
\label{fig:constraints}
\end{figure}

To generate partial labels $\mathring{\mathbf{Y}} = ( \mathring{\mathbf{Y}}_w, \mathring{\mathbf{Y}}_s, \mathring{\mathbf{Y}}_o )$ for the water, sky and obstacle class respectively, we introduce domain-specific constraints (Figure~\ref{fig:constraints}) extrapolated from (i) obstacle annotations and (ii) horizon location estimated from the IMU. The estimated horizon splits the image into two sets: regions above it ($H^\uparrow$) and regions below it ($H^\downarrow$). Similarly, water edge annotations define sets $W^\uparrow$ and $W^\downarrow$ for regions above and below, respectively, and bounding boxes define the set $O$ of rectangular regions tightly containing dynamic obstacles.

Using this notation, we can define class-restricted regions $C_c$, in which the class $c$ cannot appear. Namely, water pixels cannot appear above the horizon or water edge ($C_w = H^\uparrow \, \cup \, W^\uparrow$), sky pixels cannot appear below the horizon or water edge ($C_s = H^\downarrow \, \cup \, W^\downarrow$) and obstacle pixels cannot appear outside object bounding boxes, except above the water edge ($C_o = O^\mathrm{C} \setminus W^\uparrow$). We can thus set the probability for class $c$ of a pixel $i$ to 0 inside the respective restricted area
\begin{equation}
    \label{eq:constraints}
    \mathring{\mathbf{Y}}_c^i = 0; i \in C_c.
\end{equation}
The corresponding labels within the areas, in which a single possible class remains after applying the constraints, are set to 1 -- we refer to such areas as \textit{constrained}.
The remaining areas are \textit{unconstrained} since more than one class is possible and we set the labels therein to 0 for all classes. This formulation allows labels to be treated as certainties for individual classes. Areas with $\mathring{\mathbf{Y}}_w^i=\mathring{\mathbf{Y}}_s^i=\mathring{\mathbf{Y}}_o^i=0$ will effectively be ignored in the segmentation loss and not contribute to feature learning.

\begin{figure}
\centering
\includegraphics[width=\linewidth]{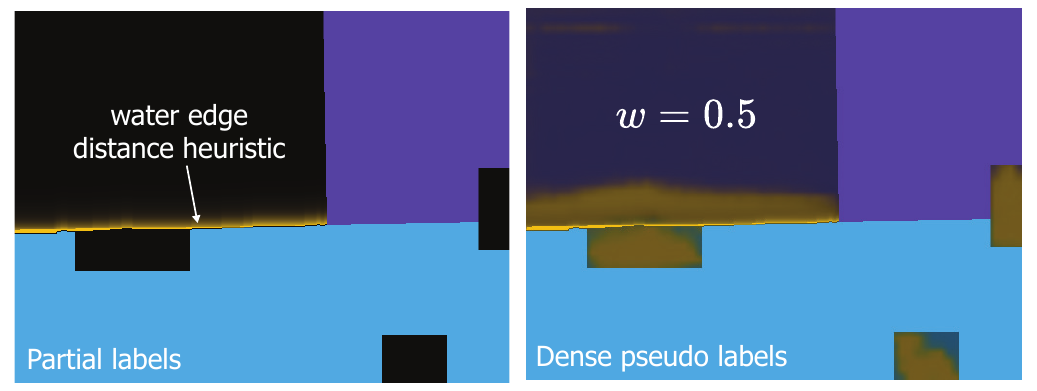}
\caption{Weak annotation constraints from Figure~\ref{fig:constraints} generate partial labels (left) with black denoting regions of unknown labels.
The labels in these regions are estimated from the warmed-up encoder features and down-weighted, producing dense pseudo labels (right).}
\label{fig:sparse_labels}
\end{figure}

From obstacle annotations, we can also infer that a pixel $i$, located immediately above the water edge, must belong to the obstacle class ($\mathring{\mathbf{Y}}_o^i = 1$). Note that with the distance from the water edge the probability decreases.
Thus we set the certainty of obstacle class for pixels above the water edge using a distance heuristic
\begin{equation}
    \mathring{\mathbf{Y}}_o^{i \in W^\uparrow} =
    \begin{cases}
        \mathrm{exp}\left(-\beta_W d_i\right) & \text{if $d_i < \theta_W$,} \\
        0 & \text{otherwise}
    \end{cases}
\end{equation}
where $d_i \geq 0$ denotes the distance of the pixel $i$ from the nearest water edge (in pixels) and $\beta_W$ is a scaling hyperparameter. To prevent learning wrong class labels in highly uncertain areas, we define a distance cutoff threshold $\theta_W$, beyond which obstacle certainty is set to 0. The resulting partial masks are visualized in Figure~\ref{fig:sparse_labels}.

\subsubsection{Training}\label{sec:warmup/training}

The model warm-up is supervised by the loss function
\begin{equation}
    \label{eq:warmup}
    L_\mathrm{wu} = L_\mathrm{foc}(\mathring{\mathbf{Y}}) + \lambda_1 L_\mathrm{pairwise} + \lambda_2 L_\mathrm{proj},
\end{equation}
where $L_\mathrm{foc}(\mathring{\mathbf{Y}})$ is a focal (segmentation) loss term~\cite{Lin2020Focal} on the generated partial labels $\mathring{\mathbf{Y}}$ (Section~\ref{sec:warmup/labels}). 

Dynamic obstacles are not uniquely labeled by constraints and cannot be learned from $\mathring{\mathbf{Y}}$. We thus leverage the bounding box annotations with a projection loss $L_\mathrm{proj}$~\cite{Tian2020BoxInst}. The projection loss provides a weak constraint on the segmentation of an obstacle, forcing the horizontal and vertical projection of the segmentation mask to match the edges of the bounding box. We apply this loss for all annotated bounding boxes. 
Further regularization is applied by using a pairwise loss~\cite{Tian2020BoxInst} $L_\mathrm{pairwise}$. This loss encourages equal labels on visually similar neighboring pixels. We adapt the pairwise loss term to a multiclass setting and apply it over the entire image (instead of per-object) so that it also supervises ambiguous areas outside dynamic obstacle annotations (\eg sky-obstacle boundary). We use the parameters from \cite{Tian2020BoxInst}.

\subsection{Estimating dense pseudo labels from features}\label{sec:dense_labels}

In this step, we estimate labels of unconstrained regions in the partial labels from encoder features of the model from the warm-up phase (Section~\ref{sec:warmup}) producing dense pseudo labels $\tilde{\mathbf{Y}}$. This process is based on the assumption, that locations corresponding to the same class cluster together in the feature latent space. Let $\hat{\mathbf{Y}} = ( \hat{\mathbf{Y}}_w, \hat{\mathbf{Y}}_s, \hat{\mathbf{Y}}_o ) \in [0,1]^{W \times H \times 3}$ be the model predictions (probabilities) for the water, sky and obstacle class and $\mathbf{F} \in \mathcal{R}^{W \times H \times C}$ be the feature maps produced by the encoder for an input image $\mathbf{I}$. Let $r(\cdot)$ be a function that corrects (\ie consolidates) the labels according to the constraints -- probabilities for restricted classes are set to 0 as in (\ref{eq:constraints}). 

From the consolidated predictions $\mathbf{R} = r(\hat{\mathbf{Y}})$, the prototype $\mathbf{p}_c$ for the class $c$ is computed as a masked average pooling over the features
\begin{equation}
    \mathbf{p}_c = \frac{\sum_{i \in \mathbf{I}}{\mathbf{R}_c^i \mathbf{F}^i}}{\sum_{i \in \mathbf{I}}{\mathbf{R}_c^i}},
\end{equation}
where $\mathbf{F}^i$ and $\mathbf{R}_c^i$ denote the features and consolidated probabilities of class $c$ at an image location $i$. Because dynamic obstacle appearance might vary greatly across instances, we use separate prototypes $\mathbf{p}_{\mathit{d}_1}, ..., \mathbf{p}_{\mathit{d}_N}$ for individual dynamic obstacles in the image (according to the annotations) and a separate single prototype $\mathbf{p}_\mathit{st}$ for all the remaining static obstacles. Two prototypes, $\mathbf{p}_w$ and $\mathbf{p}_s$, are extracted for the water and sky class respectively. 

To re-estimate the class labels based on the features, we first compute feature similarity with the prototypes at every image location to produce similarity maps $\mathbf{S}$ for each of the prototypes. The similarity map $\mathbf{S}_c$ for class $c$ at image location $i$ is computed by cosine similarity
\begin{equation}
    \mathbf{S}_c^i = \frac{\mathbf{F}^i \cdot \mathbf{p}_c}{\left \| \mathbf{F}^i \right \| \left \| \mathbf{p}_c \right \|}.
\end{equation}
Obstacle similarity maps are then merged into $\mathbf{S}_o$ as follows: similarity maps for each of the dynamic obstacle ($\mathbf{S}_{d_1}, ..., \mathbf{S}_{d_N}$) are applied inside their respective obstacle bounding boxes, and the static obstacle similarity map $\mathbf{S}_{\mathit{st}}$ is used elsewhere. In areas where annotations of multiple dynamic obstacles overlap, the maximum of their respective similarities is used. 

The per-class probabilities at each position $i$ are computed as a softmax over the class similarities, \ie
\begin{equation}
    \tilde{\mathbf{P}}_c^i = \frac{\mathrm{exp}(\beta_S \, \mathbf{S}_c^i)}{\sum_k{\mathrm{exp}(\beta_S \, \mathbf{S}_k^i)}},
\end{equation}
where $\mathbf{S} = \left ( \mathbf{S}_w, \mathbf{S}_s, \mathbf{S}_o \right )$ are the class similarity maps and $\beta_S$ is a scaling hyperparameter. The resulting probabilities are consolidated with the domain-specific constraints (Section \ref{sec:warmup/labels}) to remove impossible configurations. 

Finally, the obtained per-class probabilities are used as most likely estimates of the missing labels in the unconstrained areas, yielding $\mathring{\mathbf{Y}}$.
Additionally, the labels in the unconstrained areas are scaled by $w<1$ to give less weight to uncertain labels. 
The final dense pseudo labels (Figure~\ref{fig:sparse_labels}) at a position $i$ are thus defined as
\begin{equation}
    \tilde{\mathbf{Y}}^i = \begin{cases}
        w \, r(\tilde{\mathbf{P}}^i) & \text{if $i$ unconstrained,}\\
        \mathring{\mathbf{Y}}^i & \text{if $i$ constrained.}
    \end{cases}
\end{equation}

\subsection{Re-training with dense pseudo labels} \label{sec:retraining}

In the final step, the segmentation network is re-trained from scratch using the dense pseudo labels $\tilde{\mathbf{Y}}$. Learning is supervised by the loss
\begin{equation}
    L_\mathrm{rt} = L_\mathrm{foc}(\tilde{\mathbf{Y}}) + \lambda_1 L_\mathrm{pairwise} + \lambda_3 L_\mathrm{ws},
\end{equation}
where $L_\mathrm{foc}(\tilde{\mathbf{Y}})$ is a focal loss on the dense pseudo labels, $L_\mathrm{pairwise}$ is the pairwise loss term defined in Section~\ref{sec:warmup/training} and $L_\mathrm{ws}$ is the WaSR water separation loss~\cite{Bovcon2020WaSR}. The projection loss term is omitted since the dynamic obstacle segmentation is now supervised through the generated dense pseudo labels instead.



\section{Results}\label{sec:experiments}

\begin{table*}
\caption{Ablation study results in terms of obstacle detection performance and water-edge accuracy (A$_W$). $\overline{\mathbf{C}}$ : no constraint consolidation during Step 2 (dense pseudo-label estimation); $\overline{\mathbf{F}}$: features are not used in Step 2; $\overline{\mathbf{RT}}$ no model re-training (\ie without Steps 2 and 3); $\overline{\mathbf{PL}}$: no pairwise loss; $\mathbf{DW}$: model warm-up is supervised by dense ground-truth}
\label{table:ablation}
\centering
\begin{tabular}{lccccccccc}
\toprule
 & & & \multicolumn{3}{c}{Overall} & \phantom{a} & \multicolumn{3}{c}{Danger zone (\textless 15m)} \\
 \cmidrule{4-6} \cmidrule{8-10}
 & A$_W$ & & Pr & Re & F1 & & Pr & Re & F1 \\
\midrule
SLR                            &  11.5 &    &  94.4 &  92.1 &  \textbf{93.2} &     &  78.4 &  95.0 &  \textbf{85.9} \\
\midrule
SLR$_{\overline{C}}$           &  11.6 &    &  93.8 &  90.3 &  92.0 &     &  \textbf{78.5} &  92.9 &  85.1 \\
SLR$_{\overline{F}}$           &  10.2 &    &  93.3 &  90.0 &  91.6 &     &  64.7 &  93.2 &  76.4 \\
SLR$_{\overline{\mathrm{RT}}}$ &   \textbf{8.9} &    &  89.9 &  \textbf{94.1} &  91.9 &     &  30.4 &  \textbf{97.4} &  46.3 \\
SLR$_{\overline{\mathrm{PL}}}$ &  12.3 &    &  93.2 &  91.8 &  92.5 &     &  77.3 &  94.4 &  85.0 \\
SLR$_{\mathrm{DW}}$            &  11.2 &    &  \textbf{95.9} &  90.5 &  93.1 &     &  77.1 &  93.2 &  84.4 \\
\bottomrule
\end{tabular}
\end{table*}

\begin{table*}
\caption{Comparison of methods trained by the new scaffolding training regime (SLR) using weak annotations and state-of-the-art methods trained classically with dense ground truth labels. The methods trained using SLR are denoted by $(\cdot)_\mathrm{SLR}$. Performance is reported in terms of F1 score, precision (Pr), recall (Re) and water-edge accuracy (A$_W$).}
\label{table:sota-comparison}
\centering
\begin{tabular}{lccccccccc}
\toprule
 & & & \multicolumn{3}{c}{Overall} & \phantom{a} & \multicolumn{3}{c}{Danger zone (\textless 15m)} \\
 \cmidrule{4-6} \cmidrule{8-10}
 & A$_W$ & & Pr & Re & F1 & & Pr & Re & F1 \\
\midrule
RefineNet~\cite{Lin2017RefineNet} & 11 &  & 89.0 & \textbf{93.0} & 91.0 &  & 45.1 & 98.1 & 61.8 \\
DeepLabV3~\cite{Chen2018DeepLab} & \textbf{10} &  & 80.1 & 92.7 & 86.0 &  & 18.6 & \textbf{98.4} & 31.3 \\
BiSeNet~\cite{Yu2018Bisenet} & \textbf{10} &  & 90.5 & 89.9 & 90.2 &  & 53.7 & 97.0 & 69.1 \\
WaSR~\cite{Bovcon2020WaSR} & 11.0 &  & 95.5 & 89.9 & 92.6 &  & 62.8 & 92.9 & 75.0 \\
\midrule
DeepLabV3$_\mathrm{SLR}$ & 11.6 &  & 94.1 & 87.2 & 90.5 &  & 74.1 & 95.0 & 83.3 \\
WaSR$_\mathrm{SLR}$ & 11.5 &  & \textbf{94.4} & 92.1 & \textbf{93.2} &  & \textbf{78.4} & 95.0 & \textbf{85.9} \\
\bottomrule
\end{tabular}
\end{table*}

\begin{figure*}
\centering
\includegraphics[width=\linewidth]{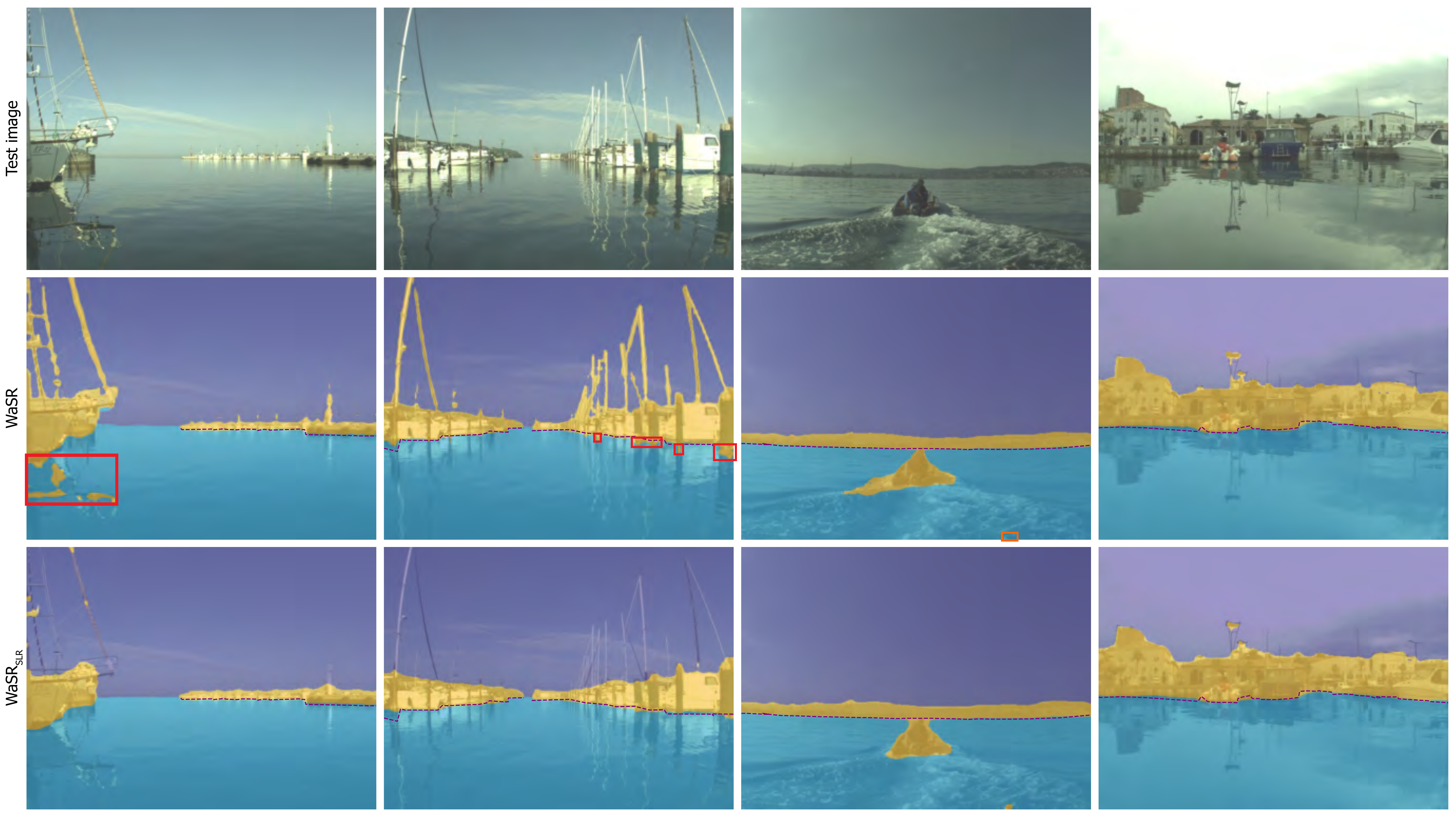}
\caption{WaSR (fully-supervised) vs. WaSR$_\mathrm{SLR}$ trained using the proposed scaffolding approach. WaSR often predicts false positive detections on water reflections of obstacles (columns 1 and 2), while WaSR$_\mathrm{SLR}$ is more robust. WaSR$_\mathrm{SLR}$ also performs well in difficult scenarios such as wakes (column 3) and heavy reflections (column 4). Best viewed on a screen.}
\label{fig:comparison}
\end{figure*}



The proposed scaffolding learning regime (SLR) is evaluated on the most recent maritime obstacle detection benchmark MODS~\cite{Bovcon2020MODS}, which contains approximately 100 annotated sequences captured under various conditions.
The evaluation protocol reflects the detection performance meaningful for practical USV navigation and separately evaluates (i) the accuracy of obstacle-water edge estimation for static obstacles and (ii) the detection performance for dynamic obstacles.

Water-edge accuracy (A$_W$) is computed as the average distance from the ground truth edge, while dynamic obstacle detection is evaluated in terms of true-positive (TP), false-positive (FP) and false-negative (FN) detections and summarized by the F1 measure, precision (Pr) and recall (Re).
A dynamic obstacle counts as detected (TP) if the coverage of the segmentation inside the ground truth bounding box is sufficient, otherwise the obstacle counts as undetected (FN). Predicted segmentations outside of the ground truth bounding boxes count as false positive detections. Detection performance is reported over the entire visible navigable area and separately only within a 15m \textit{danger zone} from the USV, where the detection performance is critical for immediate collision prevention.


The maritime obstacle dataset MaSTr1325~\cite{Bovcon2019Mastr}, which contains 1325 fully segmented images recorded by a USV, is used for training. 
To test the proposed scaffolding learning regime, the dataset was annotated with weak annotations (water edge and bounding boxes).
Following \cite{Bovcon2019Mastr}, the networks are trained in 12 image batches, with RMSProp optimizer momentum 0.9, initial learning rate $10^{-6}$, standard polynomial reduction decay 0.9 and with random image augmentations.
 

SLR is analyzed using the most recent maritime obstacle detection network WaSR~\cite{Bovcon2020WaSR}, which employs a ResNet-101 pre-trained on COCO~\cite{Lin2014COCO} as the encoder.
In warm-up and re-training phases, the decoder is randomly initialized, and features from the penultimate encoder residual block (third) are used in the dense pseudo-label estimation phase (Section~\ref{sec:dense_labels}). The focal loss parameter is set to $\gamma = 2$, while the remaining hyperparamters are set to $\lambda_1 = \lambda_2 = 1$, $\lambda_3 = 10^{-4}$, $\beta_W = 0.5$, $\theta_W = 12$, $\beta_S = 20$ and $w = 0.5$. In a preliminary study, we determined that the performance of the model is not sensitive to the exact values of the hyperparameters.
The number of training epochs in warm-up and re-training is set to 25 and 50, respectively.

\subsection{Ablation study}\label{sec:experiments/ablation}

Results of the ablation study are summarized in Table~\ref{table:ablation}. 
We first evaluate the impact of applying constraints during the dense pseudo-label estimation step (Section~\ref{sec:dense_labels}). A version that does not apply the constraints during this step (SLR$_{\overline{C}}$ ) incurs a 0.9\% and 1.3\% performance drop in water edge estimation and F1 score, respectively. Directly using model predictions as the class probabilities instead of generating similarities from features (SLR$_{\overline{F}}$) leads to a substantial detection performance drop (1.7\% overall and 11.7\% within the danger zone). This drop is mainly due to the increased number of FPs.

\begin{figure}
\centering
\includegraphics[width=\linewidth]{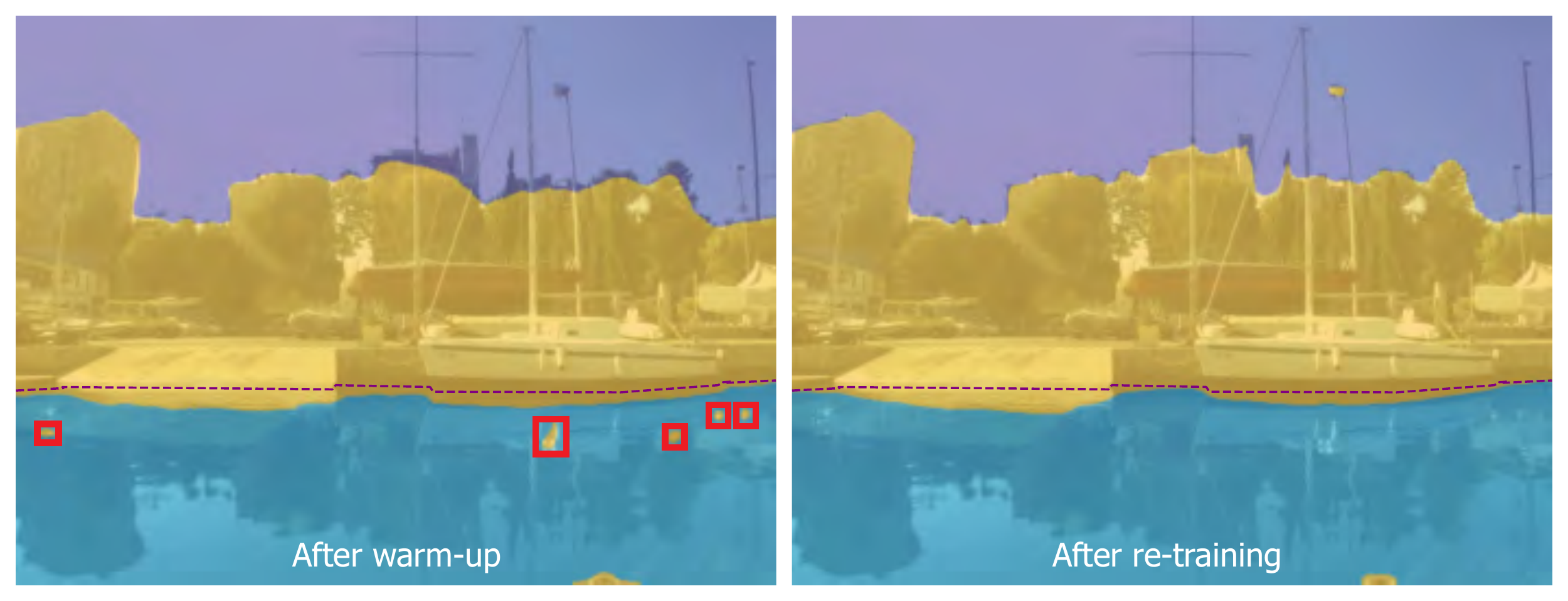}
\caption{Effects of model re-training using the estimated dense pseudo labels. After the initial warm-up phase (left), the network tends to predict FPs on unseen data (\eg reflections). 
Re-training improves performance by reducing FPs (right).}
\label{fig:introspection_comp}
\end{figure}

Next, we train only with the warm-up step and skip the re-training step (SLR$_{\overline{\mathrm{RT}}}$). For a fair comparison, the number of epochs is increased to 75 to match the total number of epochs in SLR. Skipping re-training results in 1.7\% and 59.9\% detection performance drop overall and within the danger zone, respectively. This is mainly due to a substantial increase in FPs (example in Figure~\ref{fig:introspection_comp}).

Finally, we analyze the contribution of the pairwise loss (SLR$_{\overline{\mathrm{PL}}}$). Disabling this loss incurs the smallest obstacle detection performance reduction (0.8\% F1 overall and 1.1\% inside the danger zone).


We also analyze the significance of annotation accuracy in the warm-up phase.
We thus replace the partial labels in the warm-up stage with full per-pixel ground truth labels (SLR$_{\mathrm{DW}}$). We find that better initial labels do not result in better performance but in fact cause a minor decrease (0.1\% F1 overall and 1.7\% inside the danger zone). 

\subsection{Increasing the scaffolding height}

\begin{table}
\caption{Results of increasing the network re-training iterations (including dense pseudo-label estimation) in terms of F1 calculated overall and inside the danger zone (F1$_D$), and water-edge estimation accuracy (A$_W$).}
\label{table:num_iterations}
\centering
\begin{tabular}{cccc}
\toprule
iterations & A$_W$ & F1 & F1$_D$ \\
\midrule
1      & \textbf{11.3} &  \textbf{93.5} &    86.1 \\
2      &  11.5         &  93.2          &    85.4 \\
3      &  11.8         &  93.0          &    \textbf{86.3}  \\
\bottomrule
\end{tabular}
\end{table}

SLR is composed of three steps, however, additional iterations of the last two steps might lead to further improvements. We thus analyze the models in which the dense pseudo-label estimation and network re-training (Section~\ref{sec:dense_labels} and Section~\ref{sec:retraining}) are executed multiple times. Dense pseudo-label estimation is performed based on the predictions and features of the previous model.
Results presented in Table~\ref{table:num_iterations} indicate that performance saturates after a single re-training step. We thus conclude that a single re-training after the warm-up stage is sufficient.



\subsection{Comparison with the state-of-the-art}\label{sec:experiments/sota}

\begin{table}
\caption{Segmentation accuracy of fully-supervised WaSR trained on dense labels and WaSR trained with SLR in terms of intersection over union for water, sky and obstacles (IoU$_w$, IoU$_s$,IoU$_o$) and classification accuracy (Acc).}
\label{table:segmentation}
\centering
\begin{tabular}{lcccc}
\toprule
 & IoU$_w$ & IoU$_s$ & IoU$_o$ & Acc \\
\midrule
Full supervision      & 99.7 & 99.8 & 98.0 & 99.8 \\
SLR  & 99.4 & 99.3 & 94.2 & 99.5 \\
\bottomrule
\end{tabular}
\end{table}

\begin{figure}
\centering
\includegraphics[width=\linewidth]{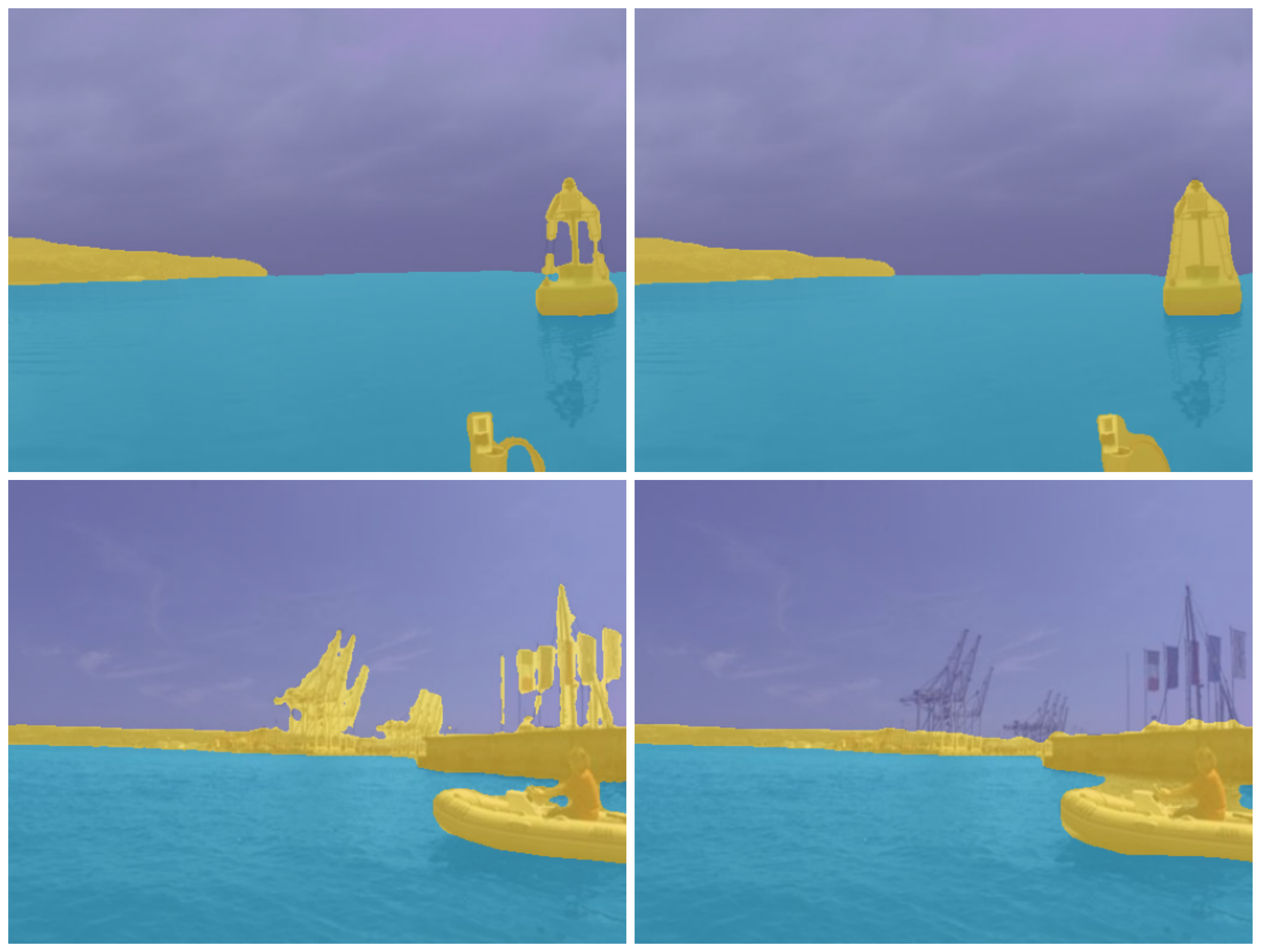}
\caption{Segmentations obtained by WaSR trained on dense ground truth (left) and with using SLR (right).}
\label{fig:segmentation}
\end{figure}


Finally, we compare the performance of the SLR-trained networks and classically-trained state-of-the-art methods from \cite{Bovcon2020MODS} (RefineNet~\cite{Lin2017RefineNet}, DeepLabV3~\cite{Chen2018DeepLab}, BiSeNet~\cite{Yu2018Bisenet} and WaSR~\cite{Bovcon2020WaSR}).
In addition to WaSR, we also demonstrate SLR on DeepLabV3 -- in the following, we refer to these two networks as WaSR$_\mathrm{SLR}$ and DeepLabV3$_\mathrm{SLR}$.
RefineNet, DeepLabV3, BiSeNet and WaSR are trained on MaSTr1325~\cite{Bovcon2019Mastr} with per-pixel ground truth labels, while DeepLabV3$_\mathrm{SLR}$ and WaSR$_\mathrm{SLR}$ are trained with weak annotations only.
Results are shown in Table~\ref{table:sota-comparison}.


Both DeepLabV3$_\mathrm{SLR}$ and WaSR$_\mathrm{SLR}$ outperform their counterparts DeepLabV3 and WaSR.  WaSR$_\mathrm{SLR}$ outperforms WaSR by 0.6\% and 13.5\% in terms of F1 score overall and inside the danger zone, respectively. Similarly, DeepLabV3$_\mathrm{SLR}$ outperforms its counterpart by 5.1\% and 90.8\% in terms of F1 score overall and inside the danger zone, respectively. The results are remarkable, given that WaSR$_\mathrm{SLR}$ and DeepLabV3$_\mathrm{SLR}$ are trained on a drastically simplified annotation scheme consisting only of obstacle bounding boxes and water edges. Note that WaSR$_\mathrm{SLR}$ outperforms all other methods and sets a new state-of-the-art on the MODS benchmark.

We find that models trained from obstacle annotations (\ie using SLR) tend to be much more robust and achieve much greater detection precision without significant loss in detection recall. For example, in DeepLabV3$_\mathrm{SLR}$ the precision inside the danger zone is increased from 18.6\% (DeepLabV3) to 74.1\% while the recall drops from 98.4\% to 95.0\% resulting in a much better overall performance. This is mainly due to the reduction in the number of false-positive detections, to which fully-supervised models seem to be much more sensitive as demonstrated in Figure~\ref{fig:comparison}. 
As expected, models trained with SLR produce slightly less accurate segmentation masks above the water edge, missing features like ship masts and thin towers (Figure~\ref{fig:comparison}). However, this does not impact obstacle avoidance as the segmentation beyond the predicted water edge is not used in the downstream task.

\subsection{Segmentation vs. detection quality}

To further compare SLR with standard fully-supervised training on dense labels, we re-trained WaSR with the two training regimes on MaSTr1325, randomly split into training (70\%) and test (30\%) images. Results are shown in \mbox{Table~\ref{table:segmentation}}. The segmentation quality of WaSR trained with SLR closely matches that of trained fully supervised, albeit at a 4\% IoU reduction on obstacles. Note that the IoU drop is due to errors incurred in regions not critical for obstacle detection (above the water edge) and slightly over-segmented objects (Figure~\ref{fig:segmentation}). This also demonstrates why IoU is not practical for measuring obstacle detection capability and is in line with observations in Figure~\ref{fig:comparison} (first two columns), where the fully-supervised network would attain a higher IoU on account of better segmenting the boat masts, yet produces more FPs and misses a small obstacle (third column) compared to the SLR-trained network.


\section{Conclusion}\label{sec:conclusion}

We proposed a novel scaffolding learning regime (SLR) for training maritime obstacle-detection-by-segmentation models using only weak annotations, which focuses on aspects important for obstacle detection. Models trained with SLR not only match the performance of fully supervised training but outperform them by a considerable margin, which is an exceptional result. Furthermore, we show that SLR is general and can be applied successfully to existing segmentation models and offers new efficient training means for future maritime obstacle detection methods.


SLR will enable the creation of large maritime obstacle detection datasets with reduced (over 20 fold) annotation effort and allow allocation of more resources to the development of better detection architectures. 
We speculate that similar methodology could also be applied to tasks such as domain transfer and beyond maritime domain to autonomous cars, using appropriate domain constraints formulations. These directions will be the topics of our future work.

\section*{Acknowledgments}
This work was supported by Slovenian research agency program P2-0214 and project J2-2506.

{\small
\bibliographystyle{ieee_fullname}
\bibliography{references}

\begin{thebibliography}{10}\itemsep=-1pt

\bibitem{Ahn2018Affinity}
Jiwoon Ahn and Suha Kwak.
\newblock {Learning Pixel-Level Semantic Affinity with Image-Level Supervision
  for Weakly Supervised Semantic Segmentation}.
\newblock In {\em Proceedings of the IEEE Computer Society Conference on
  Computer Vision and Pattern Recognition}, 2018.

\bibitem{Akiva2020Finding}
Peri Akiva, Kristin Dana, Peter Oudemans, and Michael Mars.
\newblock {Finding berries: Segmentation and counting of cranberries using
  point supervision and shape priors}.
\newblock In {\em IEEE Computer Society Conference on Computer Vision and
  Pattern Recognition Workshops}, volume 2020-June, 2020.

\bibitem{Bearman2016What}
Amy Bearman, Olga Russakovsky, Vittorio Ferrari, and Li Fei-Fei.
\newblock {What’s the point: Semantic segmentation with point supervision}.
\newblock In {\em Lecture Notes in Computer Science (including subseries
  Lecture Notes in Artificial Intelligence and Lecture Notes in
  Bioinformatics)}, volume 9911 LNCS, 2016.

\bibitem{Bhat2020Learning}
Goutam Bhat, Felix~Järemo Lawin, Martin Danelljan, Andreas Robinson, Michael
  Felsberg, Luc Van~Gool, and Radu Timofte.
\newblock {Learning What to Learn for Video Object Segmentation}.
\newblock In {\em European Conference on Computer Vision}, 3 2020.

\bibitem{Bovcon2018Obstacle}
Borja Bovcon and Matej Kristan.
\newblock {Obstacle Detection for USVs by Joint Stereo-View Semantic
  Segmentation}.
\newblock In {\em IEEE International Conference on Intelligent Robots and
  Systems}, 2018.

\bibitem{Bovcon2020WaSR}
Borja Bovcon and Matej Kristan.
\newblock {A water-obstacle separation and refinement network for unmanned
  surface vehicles}.
\newblock In {\em Proceedings of the IEEE International Conference on Robotics
  and Automation}, pages 9470--9476, 2020.

\bibitem{Bovcon2018Stereo}
Borja Bovcon, Rok Mandeljc, Janez Per{\v{s}}, and Matej Kristan.
\newblock {Stereo obstacle detection for unmanned surface vehicles by
  IMU-assisted semantic segmentation}.
\newblock {\em Robotics and Autonomous Systems}, 104, 2018.

\bibitem{Bovcon2019Mastr}
Borja Bovcon, Jon Muhovi{\v{c}}, Janez Per{\v{s}}, and Matej Kristan.
\newblock {The MaSTr1325 dataset for training deep USV obstacle detection
  models}.
\newblock In {\em 2019 IEEE/RSJ International Conference on Intelligent Robots
  and Systems (IROS)}, pages 3431--3438, 2019.

\bibitem{Bovcon2020MODS}
Borja Bovcon, Jon Muhovi{\v{c}}, Duško Vranac, Dean Mozeti{\v{c}}, Janez
  Per{\v{s}}, and Matej Kristan.
\newblock {MODS -- A USV-oriented object detection and obstacle segmentation
  benchmark}.
\newblock 5 2021.

\bibitem{Cane2019Evaluating}
Tom Cane and James Ferryman.
\newblock {Evaluating deep semantic segmentation networks for object detection
  in maritime surveillance}.
\newblock In {\em Proceedings of AVSS 2018 - 2018 15th IEEE International
  Conference on Advanced Video and Signal-Based Surveillance}, 2019.

\bibitem{Chen2018DeepLab}
Liang~Chieh Chen, George Papandreou, Iasonas Kokkinos, Kevin Murphy, and
  Alan~L. Yuille.
\newblock {DeepLab: Semantic Image Segmentation with Deep Convolutional Nets,
  Atrous Convolution, and Fully Connected CRFs}.
\newblock {\em IEEE Transactions on Pattern Analysis and Machine Intelligence},
  40(4), 2018.

\bibitem{Chen2017Rethinking}
Liang~Chieh Chen, George Papandreou, Florian Schroff, and Hartwig Adam.
\newblock {Rethinking atrous convolution for semantic image segmentation}, 6
  2017.

\bibitem{Dai2015BoxSup}
Jifeng Dai, Kaiming He, and Jian Sun.
\newblock {BoxSup: Exploiting Bounding Boxes to Supervise Convolutional
  Networks for Semantic Segmentation}.
\newblock In {\em Proceedings of the IEEE International Conference on Computer
  Vision (ICCV)}, 12 2015.

\bibitem{He2020Mask}
Kaiming He, Georgia Gkioxari, Piotr Doll{\'{a}}r, and Ross Girshick.
\newblock {Mask R-CNN}.
\newblock {\em IEEE Transactions on Pattern Analysis and Machine Intelligence},
  42(2):386--397, 2 2020.

\bibitem{Hsu2019Weakly}
Cheng~Chun Hsu, Kuang~Jui Hsu, Chung~Chi Tsai, Yen~Yu Lin, and Yung~Yu Chuang.
\newblock {Weakly supervised instance segmentation using the bounding box
  tightness prior}.
\newblock In {\em Advances in Neural Information Processing Systems},
  volume~32, 2019.

\bibitem{Huang2018Seeded}
Zilong Huang, Xinggang Wang, Jiasi Wang, Wenyu Liu, and Jingdong Wang.
\newblock {Weakly-Supervised Semantic Segmentation Network with Deep Seeded
  Region Growing}.
\newblock In {\em Proceedings of the IEEE Computer Society Conference on
  Computer Vision and Pattern Recognition}, 2018.

\bibitem{Hung2019Adversarial}
Wei~Chih Hung, Yi~Hsuan Tsai, Yan~Ting Liou, Yen~Yu Lin, and Ming~Hsuan Yang.
\newblock {Adversarial learning for semi-supervised semantic segmentation}.
\newblock In {\em British Machine Vision Conference 2018, BMVC 2018}, 2019.

\bibitem{Kalluri2019Universal}
Tarun Kalluri, Girish Varma, Manmohan Chandraker, and C.~V. Jawahar.
\newblock {Universal semi-supervised semantic segmentation}.
\newblock In {\em Proceedings of the IEEE International Conference on Computer
  Vision}, volume 2019-Octob, 2019.

\bibitem{Khoreva2017Simple}
Anna Khoreva, Rodrigo Benenson, Jan Hosang, Matthias Hein, and Bernt Schiele.
\newblock {Simple does It: Weakly supervised instance and semantic
  segmentation}.
\newblock In {\em Proceedings - 30th IEEE Conference on Computer Vision and
  Pattern Recognition, CVPR 2017}, volume 2017-Janua, 2017.

\bibitem{Kim2019Vision}
Hanguen Kim, Jungmo Koo, Donghoon Kim, Byeolteo Park, Yonggil Jo, Hyun Myung,
  and Donghwa Lee.
\newblock {Vision-Based Real-Time Obstacle Segmentation Algorithm for
  Autonomous Surface Vehicle}.
\newblock {\em IEEE Access}, 7, 2019.

\bibitem{Kristan2016Fast}
Matej Kristan, Vildana~Sulić Kenk, Stanislav Kova{\v{c}}i{\v{c}}, and Janez
  Per{\v{s}}.
\newblock {Fast Image-Based Obstacle Detection from Unmanned Surface Vehicles}.
\newblock {\em IEEE Transactions on Cybernetics}, 46(3), 2016.

\bibitem{Kulharia2020Box2Seg}
Viveka Kulharia, Siddhartha Chandra, Amit Agrawal, Philip Torr, and Ambrish
  Tyagi.
\newblock {Box2Seg: Attention Weighted Loss and Discriminative Feature Learning
  for Weakly Supervised Segmentation}.
\newblock In {\em Lecture Notes in Computer Science (including subseries
  Lecture Notes in Artificial Intelligence and Lecture Notes in
  Bioinformatics)}, volume 12372 LNCS, 2020.

\bibitem{Lee2018Image}
Sung~Jun Lee, Myung~Il Roh, Hye~Won Lee, Ji~Sang Ha, and Il~Guk Woo.
\newblock {Image-based ship detection and classification for unmanned surface
  vehicle using real-time object detection neural networks}.
\newblock In {\em Proceedings of the International Offshore and Polar
  Engineering Conference}, volume 2018-June, 2018.

\bibitem{Lin2016ScribbleSup}
Di Lin, Jifeng Dai, Jiaya Jia, Kaiming He, and Jian Sun.
\newblock {ScribbleSup: Scribble-Supervised Convolutional Networks for Semantic
  Segmentation}.
\newblock {\em Proceedings of the IEEE Computer Society Conference on Computer
  Vision and Pattern Recognition}, 2016-Decem:3159--3167, 4 2016.

\bibitem{Lin2017RefineNet}
Guosheng Lin, Anton Milan, Chunhua Shen, and Ian Reid.
\newblock {RefineNet: Multi-path refinement networks for high-resolution
  semantic segmentation}.
\newblock In {\em Proceedings - 30th IEEE Conference on Computer Vision and
  Pattern Recognition, CVPR 2017}, volume 2017-Janua, 2017.

\bibitem{Lin2020Focal}
Tsung~Yi Lin, Priya Goyal, Ross Girshick, Kaiming He, and Piotr Dollar.
\newblock {Focal Loss for Dense Object Detection}.
\newblock {\em IEEE Transactions on Pattern Analysis and Machine Intelligence},
  42(2), 2020.

\bibitem{Lin2014COCO}
Tsung~Yi Lin, Michael Maire, Serge Belongie, James Hays, Pietro Perona, Deva
  Ramanan, Piotr Doll{\'{a}}r, and C.~Lawrence Zitnick.
\newblock {Microsoft COCO: Common objects in context}.
\newblock In {\em Lecture Notes in Computer Science (including subseries
  Lecture Notes in Artificial Intelligence and Lecture Notes in
  Bioinformatics)}, volume 8693 LNCS, pages 740--755. Springer Verlag, 5 2014.

\bibitem{Ma2020Convolutional}
Liyong Ma, Wei Xie, and Haibin Huang.
\newblock {Convolutional neural network based obstacle detection for unmanned
  surface vehicle}.
\newblock {\em Mathematical Biosciences and Engineering}, 17(1), 2020.

\bibitem{Maninis2018Deep}
K.~K. Maninis, S. Caelles, J. Pont-Tuset, and L. Van~Gool.
\newblock {Deep Extreme Cut: From Extreme Points to Object Segmentation}.
\newblock In {\em Proceedings of the IEEE Computer Society Conference on
  Computer Vision and Pattern Recognition}, 2018.

\bibitem{Mittal2021Semi}
Sudhanshu Mittal, Maxim Tatarchenko, and Thomas Brox.
\newblock {Semi-Supervised Semantic Segmentation with High- And Low-Level
  Consistency}.
\newblock {\em IEEE Transactions on Pattern Analysis and Machine Intelligence},
  43(4), 2021.

\bibitem{Moosbauer2019Benchmark}
Sebastian Moosbauer, Daniel Konig, Jens Jakel, and Michael Teutsch.
\newblock {A benchmark for deep learning based object detection in maritime
  environments}.
\newblock In {\em IEEE Computer Society Conference on Computer Vision and
  Pattern Recognition Workshops}, volume 2019-June, pages 916--925, 2019.

\bibitem{Patino2017}
Luis Patino, Tahir Nawaz, Tom Cane, and James Ferryman.
\newblock {PETS 2017: Dataset and Challenge}.
\newblock In {\em IEEE Computer Society Conference on Computer Vision and
  Pattern Recognition Workshops}, volume 2017-July, pages 2126--2132, 2017.

\bibitem{Prasad2017Video}
Dilip~K. Prasad, Deepu Rajan, Lily Rachmawati, Eshan Rajabally, and Chai Quek.
\newblock {Video Processing From Electro-Optical Sensors for Object Detection
  and Tracking in a Maritime Environment: A Survey}.
\newblock {\em IEEE Transactions on Intelligent Transportation Systems}, 18(8),
  2017.

\bibitem{Ren2017Faster}
Shaoqing Ren, Kaiming He, Ross Girshick, and Jian Sun.
\newblock {Faster R-CNN: Towards Real-Time Object Detection with Region
  Proposal Networks}.
\newblock {\em IEEE Transactions on Pattern Analysis and Machine Intelligence},
  39(6):1137--1149, 6 2017.

\bibitem{Souly2017Semi}
Nasim Souly, Concetto Spampinato, and Mubarak Shah.
\newblock {Semi Supervised Semantic Segmentation Using Generative Adversarial
  Network}.
\newblock In {\em Proceedings of the IEEE International Conference on Computer
  Vision}, volume 2017-Octob, 2017.

\bibitem{Steccanella2020}
L. Steccanella, D.~D. Bloisi, A. Castellini, and A. Farinelli.
\newblock {Waterline and obstacle detection in images from low-cost autonomous
  boats for environmental monitoring}.
\newblock {\em Robotics and Autonomous Systems}, 124, 2020.

\bibitem{Tian2020BoxInst}
Zhi Tian, Chunhua Shen, Xinlong Wang, and Hao Chen.
\newblock {BoxInst: High-Performance Instance Segmentation with Box
  Annotations}.
\newblock 2020.

\bibitem{Vernaza2017Learning}
Paul Vernaza and Manmohan Chandraker.
\newblock {Learning random-walk label propagation for weakly-supervised
  semantic segmentation}.
\newblock In {\em Proceedings - 30th IEEE Conference on Computer Vision and
  Pattern Recognition, CVPR 2017}, volume 2017-Janua, 2017.

\bibitem{Wang2020Self}
Yude Wang, Jie Zhang, Meina Kan, Shiguang Shan, and Xilin Chen.
\newblock {Self-Supervised Equivariant Attention Mechanism for Weakly
  Supervised Semantic Segmentation}.
\newblock In {\em Proceedings of the IEEE Computer Society Conference on
  Computer Vision and Pattern Recognition}, 2020.

\bibitem{Wei2018Dilated}
Yunchao Wei, Huaxin Xiao, Honghui Shi, Zequn Jie, Jiashi Feng, and Thomas~S.
  Huang.
\newblock {Revisiting Dilated Convolution: A Simple Approach for Weakly- and
  Semi-Supervised Semantic Segmentation}.
\newblock In {\em Proceedings of the IEEE Computer Society Conference on
  Computer Vision and Pattern Recognition}, 2018.

\bibitem{Yang2019Surface}
Jie Yang, Yinghao Li, Qingnian Zhang, and Yongmei Ren.
\newblock {Surface Vehicle Detection and Tracking with Deep Learning and
  Appearance Feature}.
\newblock In {\em 2019 5th International Conference on Control, Automation and
  Robotics, ICCAR 2019}, 2019.

\bibitem{Yu2018Bisenet}
Changqian Yu, Jingbo Wang, Chao Peng, Changxin Gao, Gang Yu, and Nong Sang.
\newblock {BiSeNet: Bilateral segmentation network for real-time semantic
  segmentation}.
\newblock In {\em Lecture Notes in Computer Science (including subseries
  Lecture Notes in Artificial Intelligence and Lecture Notes in
  Bioinformatics)}, volume 11217 LNCS, pages 334--349, 8 2018.

\bibitem{Zhang2020Weakly}
Jing Zhang, Xin Yu, Aixuan Li, Peipei Song, Bowen Liu, and Yuchao Dai.
\newblock {Weakly-Supervised Salient Object Detection via Scribble
  Annotations}.
\newblock In {\em Proceedings of the IEEE Computer Society Conference on
  Computer Vision and Pattern Recognition}, 2020.

\bibitem{Zhao2021Generating}
Bin Zhao, Goutam Bhat, Martin Danelljan, Luc Van~Gool, and Radu Timofte.
\newblock {Generating Masks from Boxes by Mining Spatio-Temporal Consistencies
  in Videos}.
\newblock 1 2021.

\bibitem{Zhao2017Pyramid}
Hengshuang Zhao, Jianping Shi, Xiaojuan Qi, Xiaogang Wang, and Jiaya Jia.
\newblock {Pyramid scene parsing network}.
\newblock In {\em Proceedings - 30th IEEE Conference on Computer Vision and
  Pattern Recognition, CVPR 2017}, volume 2017-Janua, pages 6230--6239, 2017.

\end{thebibliography}
}

\end{document}